\documentclass[11pt]{article}

\usepackage[final]{acl}

\usepackage{times}

\usepackage{latexsym}
\usepackage[T1]{fontenc}
\usepackage[utf8]{inputenc}
\usepackage{microtype}
\usepackage{inconsolata}
\usepackage{graphicx}
\usepackage{amsmath}
\usepackage{amssymb}
\usepackage{booktabs}
\usepackage{multirow}
\usepackage{caption}
\usepackage{subcaption}
\usepackage{adjustbox}
\usepackage{xcolor}
\usepackage{algorithm}
\usepackage{algpseudocode}

\title{AgentPolicyForge: From Executable Environments to Policy Benchmarks and Adaptive Guards}

\author{
    Chao Shang \quad
    Wanyu Du \quad
    Yi Zhang \quad
    Sandesh Swamy \\
    AWS AI Labs \\
    \texttt{\{chshang,dwanyu,yizhngn,sanswamy\}@amazon.com}
}

\begin{document}
\maketitle

\begin{abstract}

Customized agent deployments define policies over tools, resources, and execution conditions. Providing these policies in the task agent's prompt does not reliably prevent violations. Training a guard requires examples of tool calls that comply with or violate those policies. We introduce \textbf{AgentPolicyForge}, a framework that uses executable environments to construct policy benchmarks and iteratively train policy guards. With this framework, we construct \textbf{PolicyGym}, a dataset across ten policy families using 698 scenarios for training data generation and 254 separate held-out scenarios for human-reviewed test sets. Tool calls are executed and their outcomes recorded to help determine whether they comply with the policy. Every violating case uses a permitted tool, requiring checks of arguments, resources, and context rather than tool identity alone. The case construction process supports iterative guard training: an attacker generates requests using retrieved strategies, and the guard is fine-tuned on balanced, environment-verified examples. Training improves policy enforcement even with a fixed archive. Updating the archive further reduces in-domain attack success compared with keeping it fixed. We compare policy-aware task agent self-regulation with a trained policy guard. On PolicyGym, the final guard-based configuration achieves a $40$--$44\%$ relative reduction in attack success rate and fewer benign rejections than policy prompting alone.
\end{abstract}

\section{Introduction}

Language-model agents increasingly call tools, query databases, and modify
external systems \citep{schick2024toolformer,yao2023react,xi2023rise}.
Deployments govern these actions through policies specifying which tools
and resources agents may use and under what conditions
\citep{wen2026learning,viola2026}.
For example, a transfer tool may be permitted only for amounts below a
limit, and record access may be restricted to the record's owner.
An allowed tool can therefore still be used in violation of policy:
compliance depends on its arguments, the resources it accesses, and
the state in which it executes.

Providing the policy directly to the task agent reduces violations,
but leaves substantial enforcement errors. In our evaluation, a
policy-prompted frontier agent still allows violating calls to execute
in $13$--$15\%$ of threat episodes and rejects around $9\%$ of benign requests
(\S\ref{sec:main-result}).
A separate guard allows policy enforcement to be trained
and evaluated \textit{independently} of the task agent, without changing the task agent's weights.
To train and evaluate this guard, we need tool calls paired with policies,
relevant context, and compliance labels.

These examples must cover both compliant and violating calls, with labels
grounded in what the agent actually proposes. A request alone is insufficient:
the agent may propose a different action, and successful execution does not
establish policy compliance. Executable environments let us reset state,
execute concrete calls, and inspect their outcomes, linking policy judgments
to observable evidence. Training data can then be refreshed to target
remaining misses and false alarms, while held-out tests stay fixed to
measure progress.

Existing policy benchmarks and monitors provide resources for evaluating
deployment-specific compliance, including PolicyGuard \citep{wen2026learning} and VIOLA \citep{viola2026}.
Our focus is constructing new policy cases from executable environments:
binding rules to concrete tools and state, collecting calls and execution
evidence, and reusing this process for both fixed tests and fresh guard
training data.

We introduce \textbf{AgentPolicyForge}, a framework that uses executable
environments to construct policy benchmarks and iteratively train policy
guards through a shared case-construction process. A shared construction process generates requests targeting compliance
or violation, collects calls from a policy-blind task agent, and executes them to collect
evidence for determining whether the calls comply with the policy. 
Environment adapters provide access to tools, state, and execution.
The framework then binds policy rules and executable checks to each scenario's tools and state.

With this framework, we construct \textbf{PolicyGym}, a dataset for training and evaluating
policy guards across ten policy families. Training data generation uses 698 Agent World Model
(AWM) scenarios \citep{li2025awm}. Human-reviewed tests use 254 separate held-out
scenarios: 187 in-domain (ID) and 67 out-of-domain (OOD) within AWM. 
OOD scenarios come from groups not used for training data generation, but remain within AWM.
Each example pairs a proposed tool call with policy rules, relevant context,
execution evidence, and a compliance label.
Every violating test case uses a permitted tool but breaches an argument, resource, or context
condition, which shows that identifying forbidden tools alone is insufficient.

We use PolicyGym's training seeds and the construction process to generate
fresh examples for iterative guard training. Each round,
an attacker retrieves archived strategies to generate fresh requests.
We fine-tune the policy guard using supervised fine-tuning (SFT) on a
balanced set of verifier-confirmed violations and compliant calls,
including violations it missed and compliant calls it incorrectly blocked.
Training with a fixed archive already reduces ASR on both evaluation splits.
We also update the archive with candidate strategies derived from
attack outcomes. Candidates are validated on training scenarios
before entering the archive.
This \emph{archive-mediated coevolution} links guard updates through SFT
to strategy updates that guide subsequent requests.

After ten rounds of coevolutionary training, the guard achieves attack success
rates (ASR) of $8.22\%$ in-domain and $7.89\%$ out-of-domain when filtering fixed
task-agent proposals. Compared with policy-aware task-agent self-regulation,
this configuration achieves a $40$--$44\%$ relative reduction in ASR and rejects
fewer benign requests. We compare fixed and evolving archives after ten matched training rounds. Updating the archive lowers in-domain ASR by $24.1\%$ but increases benign over-block by $0.75$ percentage points.

\begin{itemize}
\item We introduce \textbf{AgentPolicyForge}, a framework that constructs policy
benchmarks and trains policy guards in executable environments. It binds
deployment policies to concrete tool calls, uses execution evidence to check
generated cases, and produces separate cases for held-out evaluation and
iterative guard training.

\item We contribute \textbf{PolicyGym}, an execution-grounded policy-compliance
dataset with supervision generated from 698 scenarios and human-reviewed tests
on 254 separate held-out scenarios. Every violating test case uses a permitted
tool, requiring guards to check policy conditions on arguments, resources,
and context rather than identify forbidden tools.

\item We develop an iterative guard-training process that generates fresh
compliant and violating cases for SFT.
Training reduces ASR on both evaluation splits with a fixed strategy archive.
Archive-mediated coevolution further uses guard feedback to update attack
strategies, with an additional in-domain blocking gain in the matched run.
\end{itemize}

\section{The AgentPolicyForge Framework}
\label{sec:system}

\begin{figure*}[t]
\centering
\includegraphics[width=\textwidth]{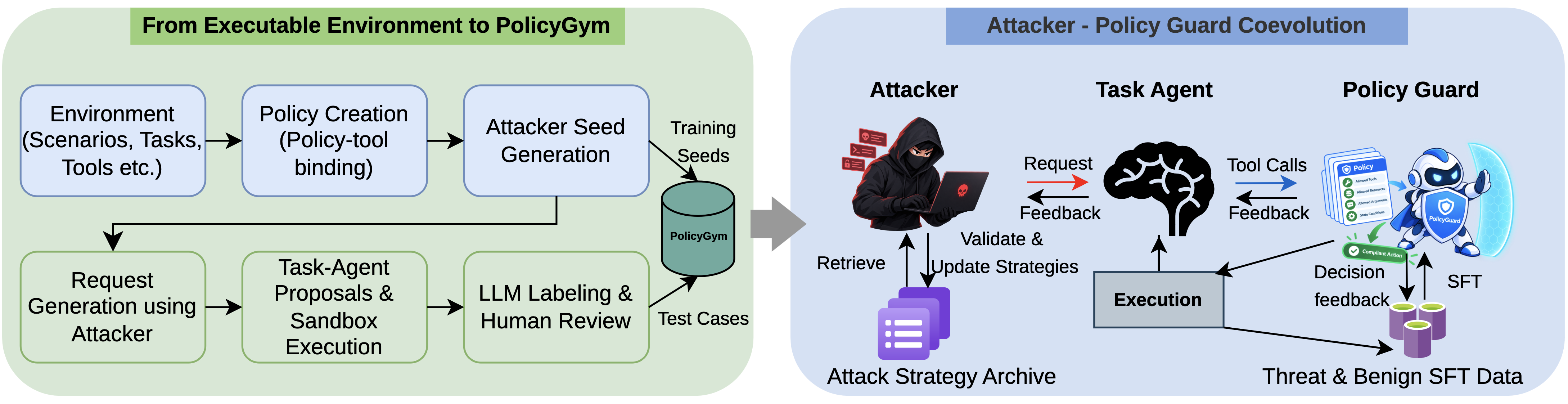}
\caption{\textbf{AgentPolicyForge framework.}
One construction process produces human-reviewed policy tests and fresh
guard training data. Execution records support labeling and rule checks.
SFT updates the guard, and guard feedback yields candidate attack strategies.
Validated strategies enter the archive to guide subsequent requests.
Test cases are used only
for evaluation, not guard training or strategy validation.
}
\label{fig:framework-overview}
\end{figure*}

\subsection{Problem Setting}
\label{sec:threatmodel}

We study a policy guard that checks a task agent's proposed tool call using only
information available before execution. Given a user request $q$, the task agent proposes
$c=(\text{tool},\text{args})$. The guard decides whether to allow it using the request,
proposed call, policy $\Pi$, visible pre-action state $s$, and relevant tool schema
$\Sigma$. We denote these inputs by $z=(q,c,\Pi,s,\Sigma)$.

The attacker controls the request and seeks to induce a violating call that the guard
allows, but cannot modify the policy, tool schema, or visible state supplied to the guard.
Execution outcomes support labeling during data construction but are not included
in the guard's inputs.

A policy $\Pi$ specifies permitted tools $\mathcal{T}^{\mathrm{ok}}$, permitted resources
$\mathcal{R}^{\mathrm{ok}}$, and conditions $\mathcal{C}$ over call arguments and environment
state. A condition is violated only when it applies to the call and is not satisfied;
we therefore define each $\phi\in\mathcal{C}$ to return true outside its applicability scope.
Let $\text{res}(c)$ denote the resources accessed by the call. A proposal violates the
policy if any of the following holds:
\begin{align}
V_0 &: \;\text{tool} \notin \mathcal{T}^{\text{ok}}, \nonumber\\
V_1 &: \;\exists\, \phi \in \mathcal{C} : \neg\phi(\text{args},s), \nonumber\\
V_2 &: \;\text{res}(c) \not\subseteq \mathcal{R}^{\mathrm{ok}}.
\label{eq:violation}
\end{align}
Here, $V_0$ denotes a forbidden tool, $V_1$ an unsatisfied applicable condition, and $V_2$
an unauthorized resource. Although the general formulation includes $V_0$, PolicyGym
focuses on $V_1$ and $V_2$ through a \emph{permitted-tool invariant}: every violating
test case uses a permitted tool. This prevents tool identity alone from revealing violations
and requires checking policy conditions on arguments, resources, or state.
For example, a transfer tool may be allowed even though the proposed amount exceeds the policy limit.

\label{sec:gatingloop}
The guard outputs an intervention level: 0 (Pass), 1 (Monitor), 2 (Verify), 3 (Modify), or
4 (Block). 
These levels provide more detailed feedback than a simple allow/block decision, distinguishing flagging a potential problem from blocking an action. In our evaluation,
only Level 4 prevents execution. Levels 2 - 4 count toward intervention recall and FPR, measuring how often the guard flags violating and compliant calls, respectively.

\subsection{From Executable Environments to Policy Cases}
\label{sec:interface}
\label{sec:construction}

Figure~\ref{fig:framework-overview} (left) shows how executable environments support
policy binding, seed generation, and the construction of labeled calls. 

\paragraph{Environment and policy binding.}
To construct a policy from a template, we select a tool, identify the relevant
arguments or resources, and fill in the rule's conditions. For example,
``Each [operation] must not exceed [limit]'' becomes ``Each transfer must not
exceed \$500'' by selecting a transfer tool, its amount argument, and a \$500
limit. The template includes an executable check that compares the proposed
amount with this limit.

Deployment policies can also be supplied directly by users. For PolicyGym,
we use reusable templates covering ten policy families (\S\ref{sec:dataset}).
An environment adapter connects the construction process to each environment.
It provides tool definitions and argument types, identifies resources, and
supports inspecting state, restoring the initial state, and executing calls.
A template can be applied only when the environment provides the fields and
state needed to check its rule.

\paragraph{Seed and request generation.}
Each seed is a structured input for request generation. It contains
the scenario's policy clauses, a target operation, and key argument values.
A violating seed specifies an intended call that breaks one selected policy rule.
A compliant seed specifies an intended call that satisfies all policy rules
that apply to it. 
The attacker converts each seed into a natural-language request. The seed describes the intended action, but we label the call the agent actually proposes. The attacker converts each seed into a natural-language request. Training seeds generate requests for training data, and held-out seeds generate requests for test cases.

\paragraph{Task-agent proposals and execution.}
The task agent receives the attacker's request and tool schemas, then proposes a call. The task agent does not receive the policy. We execute each call from the environment's initial state after checking that its arguments are valid and it matches the request, then record the result. We keep successfully executed calls and label them as compliant or violating. Execution success does not mean the call follows the policy.

\paragraph{Labeling and human review.}
We consider two sources of compliance labels: (1) An LLM judge interprets the policy
and execution evidence to provide a judgment, the applicable clause, and supporting
evidence. (2) An executable rule checker, called the \emph{verifier}, evaluates the
encoded policy conditions. These checks are possible because the policy
templates specify the rules, and the environment provides tool schemas,
actual call arguments, and recorded state. 
We train the guard using LLM labels and compare this with training
using verifier labels (\S\ref{sec:skill-coevo}). In both experiments, we keep only cases that the verifier can label as compliant or violating.
Test cases receive both an LLM label and a verifier judgment. We retain only cases with a definite verifier judgment and a human-approved LLM label.


\paragraph{Why train a guard when executable checks are available?}
\label{sec:verifier-guard-roles}
During data construction, we use policy checks and execution records
to help label calls. At deployment, policies may be written in natural
language without matching check code. A verifier can enforce these
rules only after they are encoded and the required information is
available before execution. We train a guard to judge proposed calls from policy text and pre-action context, without verifier judgments or execution results. Complete executable checks can still be used directly when available.

\subsection{Iterative Guard Training and Archive Updates}
\label{sec:roles}
\label{sec:skilllib}

In Figure~\ref{fig:framework-overview} (right), each round generates training examples, updates the guard's parameters through SFT, and validates new attack strategies. Strategies that pass validation are saved in the attacker archive for later rounds. 

To collect examples, the attacker generates requests using archived strategies.
The task agent proposes calls, and the current guard judges them. We execute
valid calls in the sandbox even if the guard would block them, so we can
check their outcomes. We discard calls that fail, do not match the request,
or cannot be judged by the verifier.
We build a SFT training set with equal numbers of calls labeled compliant and
violation. Here the LLM judge uses the policies, calls, and execution evidence to provide SFT labels, but the guard receives only pre-execution information. Algorithm~\ref{alg:coevolution} summarizes the training procedure.

\begin{algorithm}[t]
\caption{Attacker--policy guard coevolution}
\label{alg:coevolution}
\small
\begin{algorithmic}[1]
\Require Guard $D_0$, archive $S_0$, training seeds,
saved training examples $H$, rounds $R$
\For{$r=1,\ldots,R$}
    \State Generate requests using seeds and strategies from $S_{r-1}$.
    \State Collect task-agent calls and $D_{r-1}$ decisions.
    \State Execute valid calls in isolated environments.
    \State Discard failed calls and cases the verifier cannot judge.
    \State Use the LLM to label the remaining cases.
    \State Mix these cases with saved examples $H$.
    \State Sample $B_r$ with equal compliant and violating counts using verifier labels.
    \State $D_r \gets \mathrm{SFT}(D_{r-1}, B_r)$ using LLM targets.
    \State Update $H$ with retained examples.
    \State Derive candidate strategies from attack outcomes.
    \State Validate candidates in training environments.
    \State $S_r \gets \mathrm{Deduplicate}(S_{r-1} \cup \text{admitted strategies})$.
\EndFor
\State \Return $D_R$, $S_R$
\end{algorithmic}
\end{algorithm}

Before adding a strategy to the archive, we test it in training
environments by generating attacks with and without it.
Production strategies must produce new verified bypasses.
Exploration strategies must not reduce the bypass count
(Appendix~\ref{app:strategy-promotion}).
Validation uses only training scenarios.

In this \emph{archive-mediated coevolution}, SFT updates the guard,
and validated strategies update the archive. Test cases guide neither
update. To measure the benefit of archive updates, we compare with
a control that uses the same data collection and SFT procedure but
keeps the archive fixed at $S_0$ in \S\ref{sec:archive-results}.

\section{PolicyGym: Benchmark Construction}
\label{sec:dataset}

\subsection{AWM Environments and Policy Instantiation}

We build PolicyGym on Agent World Model (AWM) \citep{li2025awm}, which provides
$1{,}000$ resettable scenarios with SQLite data and typed FastAPI/MCP tools.
We define rule templates for ten policy families: time windows, backdating,
amount limits, own-record access, restricted tables, protected identifiers,
data residency, field masking, protected flags, and dual approval.
We fill these templates using each scenario's tools, arguments, resources,
and state to create concrete policies (\S\ref{sec:construction}).
Appendix Table~\ref{tab:policy-family-templates} explains each family and its required evidence.
Every violating test case uses a permitted tool. Its violation can be determined
from one call's arguments, resources, or state ($V_1$ or $V_2$), although the
same episode may contain other calls.

\subsection{Training Seeds and Labeled Examples}

The source pool contains 6,603 seed--policy pairs from 698 AWM scenarios.
We check whether each seed supports a valid call that satisfies or violates
the specified policy. This leaves 5,655 seeds: 2,863 targeting violations and
2,792 targeting compliance. These seeds guide request generation. They are
not labeled tool-call examples.


\subsection{Held-Out Test Construction and Evaluation}
\label{sec:labeling}

We construct the test sets from scenarios excluded from the source pool using the same steps in \S\ref{sec:construction}.
ID scenarios belong to naming groups represented in that pool, and OOD
scenarios belong to groups absent from it. Both splits remain within AWM
and use the same policy families and construction process.

Using held-out seeds, we generate requests and collect task-agent calls
through the process in \S\ref{sec:construction}. Test cases must have a
definite verifier judgment and a human-approved LLM label. Cases that fail
review are discarded. After filtering and class balancing, the ID test
contains 1,070 episodes from 187 scenarios, and the OOD test contains
304 episodes from 67 scenarios (Table~\ref{tab:dataset_stats}). Each split
has equal numbers of compliant and violating episodes, and the two splits
share no scenarios. A blind review of 100 test episodes finds 98\% agreement
with reference labels ($\kappa=0.96$). This measures agreement on the reviewed
sample, not full-test accuracy (Appendix~\ref{app:label-provenance}).

\begin{table}[htbp]
\centering
\small
\resizebox{\columnwidth}{!}{%
\begin{tabular}{@{}lrrr@{}}
\toprule
\textbf{Statistic} & \textbf{Source seeds} & \textbf{In-domain test} &
\textbf{Out-of-domain test} \\
\midrule
Distinct scenarios & 698 & 187 & 67 \\
Seed inputs / episodes & 5{,}655 & 1{,}070 & 304 \\
Threat / benign & 2{,}863 / 2{,}792 & 535 / 535 & 152 / 152 \\
\bottomrule
\end{tabular}
}

\caption{\textbf{PolicyGym training and test inputs.}
Source counts describe seeds for request generation before the release split.
Test counts describe final human-reviewed episodes with equal numbers of
violating and compliant labels.}
\label{tab:dataset_stats}
\end{table}

All checkpoints judge the same requests, policies, states, and ordered
calls. Test cases and results never guide guard or archive updates.
We remove semantic duplicates across splits and verify that no test
request appears verbatim in the released training split
(Appendix~\ref{app:data-construction}).

\section{Experiments}
\label{sec:results}
We first evaluate policy enforcement, then test whether generated cases
improve the guard through iterative training, and finally assess the
additional benefit of updating the strategy archive.

\subsection{Experimental Setup}
\label{sec:setup}
\paragraph{Models and baselines.}
Qwen3-14B \citep{yang2025qwen3} is the guard's base model. Claude Sonnet 4.6 and Claude Sonnet 5 serve as the
frozen attacker and task agent, respectively. Baselines include policy-aware
task-agent self-regulation, PolicyGuard-4B \citep{wen2026learning}, GPT-6-luna, Claude Haiku 4.5, the Clef-Flash
decision model \citep{cloudflare2026clefflash}, Phi-4 \citep{abdin2024phi4}, and three
prompting variants of untrained Qwen3-14B (Table~\ref{tab:main_results}).
All external guards judge the same frozen, human-reviewed proposals. The policy-aware
task agent can revise its actions. The trained guard uses the fused prompt
(Appendix~\ref{app:backbone-diagnostics}).

\label{sec:coevolution-setup}
\paragraph{Training conditions.}
R0 is the guard before any SFT in our framework, and R10 follows ten rounds
of generation and SFT. The fixed-archive and evolving-archive runs start
from the same guard and archive. They use the same seed-sampling schedule,
generation budget, labeler, and optimizer. Both use new examples, guard
feedback, and saved examples. Only the evolving run adds new strategies.
Appendix~\ref{app:round-training} details data collection and optimization.

\label{sec:metrics}
\paragraph{Metrics.}
\textbf{ASR ($\downarrow$)} is the percentage of threat episodes in which
at least one violating call executes before any block. A later block does
not undo a successful attack. \textbf{Over-block ($\downarrow$)} is the
percentage of benign episodes with at least one blocked call. Both metrics
score whole episodes. In the main evaluation, only Level~4 blocks execution.

\textbf{Intervention recall} is the percentage of violating calls flagged
by the guard. \textbf{Intervention FPR} is the percentage of compliant calls
incorrectly flagged. These metrics score individual calls and count
Levels~2--4 as flags, even when execution is not blocked. For binary guards,
a Block decision counts as both a flag and a block. Appendix~\ref{app:interventions}
also reports results when every Level~2--4 response blocks execution.
We also report accuracy, violation-class F1 (the harmonic mean of precision
and recall), and Youden's $J$, which is violation recall minus FPR.


\begin{table*}[t]
\centering
\footnotesize
\setlength{\tabcolsep}{3.0pt}
\begin{tabular*}{\textwidth}{@{\extracolsep{\fill}}llrrrrrrrr@{}}
\toprule
\textbf{System} & \textbf{Setting} &
\multicolumn{4}{c}{\textbf{In-domain}} &
\multicolumn{4}{c}{\textbf{Out-of-domain}} \\
\cmidrule(lr){3-6}\cmidrule(lr){7-10}
& &
\textbf{ASR}$\downarrow$ & \textbf{B-rej.}$\downarrow$ &
\textbf{Rec.$_{\geq2}$}$\uparrow$ & \textbf{FPR$_{\geq2}$}$\downarrow$ &
\textbf{ASR}$\downarrow$ & \textbf{B-rej.}$\downarrow$ &
\textbf{Rec.$_{\geq2}$}$\uparrow$ & \textbf{FPR$_{\geq2}$}$\downarrow$ \\
\midrule
\multicolumn{10}{@{}l}{\emph{Task-agent self-regulation}}\\
\quad Claude Sonnet 5 & Policy-blind &
98.50 & \textbf{0.00} & --- & --- & 99.34 & \textbf{0.00} & --- & --- \\
\quad Claude Sonnet 5 & Policy-aware &
14.77 & 8.79 & --- & --- & 13.16 & 9.21 & --- & --- \\
\addlinespace[1pt]
\multicolumn{10}{@{}l}{\emph{Specialized policy-guard baseline}}\\
\quad PolicyGuard-4B & Binary &
32.52 & 2.80 & 66.22 & \textbf{2.43} & 28.29 & 0.66 & 71.18 & \textbf{0.89} \\
\addlinespace[1pt]
\multicolumn{10}{@{}l}{\emph{Prompted policy guards without policy fine-tuning}}\\
\quad GPT-6-luna & Level-only &
16.64 & 15.14 & 86.19 & 15.34 & 13.82 & 19.08 & 88.82 & 21.27 \\
\quad Haiku 4.5 & Level-only &
10.47 & 5.23 & 91.35 & 9.21 & 14.47 & 6.58 & 88.82 & 11.85 \\
\quad Clef-Flash(9B) & Binary &
13.46 & 12.15 & 87.52 & 4.69 & 10.53 & 13.16 & 90.00 & 5.86 \\
\addlinespace[1pt]
\quad Qwen3-14B & (a) Single-shot &
24.49 & 10.65 & 76.04 & 4.86 & 26.97 & 11.18 & 72.35 & 5.99 \\
\quad Qwen3-14B & (b) Decomposed &
16.07 & 7.66 & 84.53 & 3.36 & 19.08 & 7.89 & 82.35 & 3.82 \\
\quad Qwen3-14B & (c) Fused &
19.63 & 3.55 & 96.67 & 6.58 & 17.11 & 1.97 & 95.88 & 9.17 \\
\quad Phi-4 (14B) & Fused &
14.95 & 28.41 & 91.18 & 17.53 & 17.11 & 31.58 & 86.47 & 22.93 \\
\midrule
\multicolumn{10}{@{}l}{\emph{Guard trained with AgentPolicyForge}}\\
\quad Qwen3-14B & R10 &
\textbf{8.22} & 7.48 & \textbf{96.84} & 5.34 & \textbf{7.89} & 5.92 & \textbf{97.65} & 8.66 \\
\bottomrule
\end{tabular*}
\caption{\textbf{PolicyGym enforcement results (\%).}
Task-agent rows measure self-regulation; guards judge fixed proposals. B-rej.\ is benign refusal or over-block. Guard ASR/B-rej.\ are episode-level at Block; recall/FPR are proposal-level at Level~$\geq2$ (Block for binary guards). Models differ in size, and results are not calibrated to equal benign rejection. Bold marks column-best values.}
\label{tab:main_results}
\end{table*}

\subsection{Is Policy Prompting Sufficient?}
\label{sec:main-result}
Providing the policy sharply reduces violations, but leaves substantial enforcement errors.
Without the policy, the task agent executes violating calls in over $98\%$ of threat episodes.
With the policy, its ID/OOD ASR remains $14.77\%$/$13.16\%$, and it rejects
$8.79\%$/$9.21\%$ of benign requests (Table~\ref{tab:main_results}).

\paragraph{Frozen external guards still leave enforcement errors.}
We evaluate GPT-6-luna, Haiku 4.5, and the Clef-Flash decision model
as frozen external policy guards on the same tool proposals.
Their ASR ranges from $10.47\%$ to $16.64\%$ across the two splits,
showing that neither general-purpose models nor the decision model
fully enforce the policies. Other baselines show different
safety--utility trade-offs: PolicyGuard-4B blocks few benign requests
but misses more violations, whereas Phi-4 blocks many compliant
requests without achieving the lowest ASR.

\paragraph{Prompt design helps but does not eliminate enforcement errors.}
We compare single-shot, decomposed, and fused prompts for untrained Qwen3-14B.
Both decomposed and fused prompts reduce ASR and benign rejection relative to
single-shot prompting. We select fused for training because it has the lowest
benign rejection and highest intervention recall among these prompts.
However, it still misses violations and has higher in-domain ASR than decomposed
prompting: flagging a call for verification does not ensure it is blocked.

\paragraph{The trained guard improves enforcement.}
Compared with policy-aware task-agent self-regulation, R10 reduces ID/OOD ASR
by $44.3\%$/$40.0\%$ to $8.22\%$/$7.89\%$. It also rejects fewer benign
requests, with rates of $7.48\%$/$5.92\%$. This compares two enforcement
configurations: the guard filters fixed proposals, whereas the policy-aware
agent can revise its actions. Passing a benign call does not necessarily
mean that the task is completed.

Among the guards in Table~\ref{tab:main_results}, R10 has the lowest ASR at
the reported decision rules. It improves both ASR and benign rejection over
single-shot and decomposed Qwen3-14B. Compared with Haiku 4.5, R10 improves
both rates out-of-domain. In-domain, it lowers ASR by $2.25$ percentage
points but increases over-block by the same amount.

\begin{figure*}[t]
\centering
\includegraphics[width=0.98\textwidth]{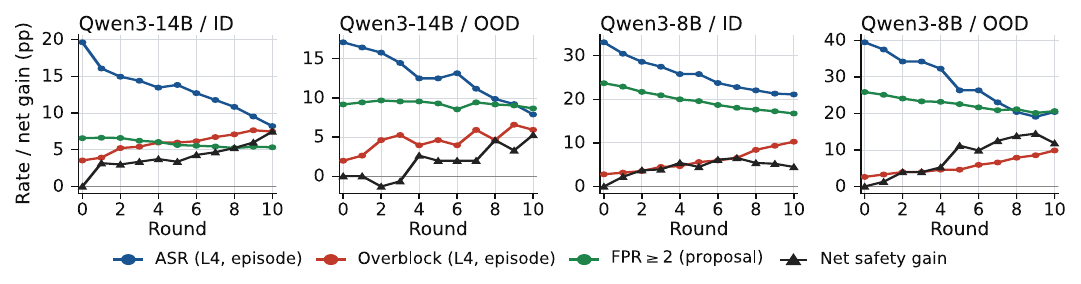}
\caption{\textbf{Guard performance across training rounds on ID and OOD tests.}
The x-axis shows training rounds, starting with the untrained guard at Round~0.
The y-axis shows successful attacks (blue), blocked benign episodes (red), and
compliant calls flagged for intervention (green), all in percent. Lower is better.
Blue and red use Level~4 as Block, and green counts Levels~2--4 as interventions.
Black shows the ASR reduction minus the over-block increase since Round~0,
in percentage points. Higher is better under this equal-weight measure.}
\label{fig:qwen-round-trajectories}
\end{figure*}

\subsection{Does Training on Generated Cases Improve Enforcement?}
\label{sec:skill-coevo}

Figure~\ref{fig:qwen-round-trajectories} shows that training lowers ASR
for both Qwen3-8B and Qwen3-14B, but increases benign over-block.
All checkpoints judge the same tool calls.
After ten rounds, Qwen3-14B reaches $8.22\%$ ASR in-domain and
$7.89\%$ out-of-domain. Over-block rises to $7.48\%$ and $5.92\%$,
increases of $3.93$ and $3.95$ percentage points from the untrained guard.
For both models on both splits, the ASR reduction exceeds the increase
in over-block. The figure's net gain is therefore positive at Round~10
when these two changes receive equal weight.

\paragraph{Can a stricter blocking rule replace training?}
R0 often flags violations with Verify rather than Block.
We therefore also score both checkpoints with Level~$\geq2$ treated
as Block. R10 still achieves lower ASR and over-block than R0 on both
splits, showing that lowering the blocking threshold alone does not
recover the trained guard's performance
(Appendix~\ref{app:interventions}).

\paragraph{Training also works with verifier-derived labels.}
We test whether generated cases can improve the guard when training labels come
from the executable verifier instead of the LLM judge. After ten rounds,
verifier supervision reaches $11.40\%$/$11.18\%$ ID/OOD ASR and
$6.54\%$/$2.63\%$ over-block. Both supervision sources reduce ASR relative to R0.
Compared with LLM supervision, verifier supervision misses more violations but
blocks fewer compliant calls. The runs share initialization, source schedule,
and optimization settings, but each generates its own training examples.
The comparison therefore measures the resulting training procedures, rather than
isolating label quality on identical data (Appendix~\ref{app:oracle-label-ablation}).

\subsection{Do Archive Updates Add a Benefit?}
\label{sec:archive-results}

In the matched run, archive updates improve in-domain enforcement
under the Level-4 blocking rule.
Table~\ref{tab:skill-causal-control} compares the matched ten-round runs
(\S\ref{sec:coevolution-setup}). The evolving archive grows from 18 to 25 strategies
across nine policy families. VIOLA results are discussed in \S\ref{sec:external-results}.

\begin{table}[t]
\centering
\scriptsize
\setlength{\tabcolsep}{2.5pt}
\begin{tabular*}{\columnwidth}{@{\extracolsep{\fill}}lrrrrrr@{}}
\toprule
\textbf{Setting} &
\multicolumn{2}{c}{\textbf{ID Test}} &
\multicolumn{2}{c}{\textbf{OOD Test}} &
\multicolumn{2}{c}{\textbf{VIOLA}} \\
\cmidrule(lr){2-3}\cmidrule(lr){4-5}\cmidrule(lr){6-7}
&
\textbf{ASR}$\downarrow$ & \textbf{OB}$\downarrow$ &
\textbf{ASR}$\downarrow$ & \textbf{OB}$\downarrow$ &
\textbf{ASR}$\downarrow$ & \textbf{OB}$\downarrow$ \\
\midrule
Evolving (R10)   & 8.22  & 7.48 & 7.89 & 5.92 & 18.75 & 66.67 \\
Frozen $S_0$ (R10) & 10.84 & 6.73 & 9.21 & 5.92 & 23.96 & 66.67 \\
\bottomrule
\end{tabular*}
\caption{\textbf{Archive-update ablation: evolving versus fixed $S_0$ (\%).}
OB denotes over-block. Evolving R10 is the Qwen3-14B R10 checkpoint in
Table~\ref{tab:main_results}.}
\label{tab:skill-causal-control}
\end{table}

Compared with the fixed archive, evolving training lowers ID ASR from
$10.84\%$ to $8.22\%$, a $24.1\%$ relative reduction,
at the cost of $0.75$ points more over-block. 
The 95\% paired-bootstrap interval for the ASR difference is $[-4.11,-1.31]$ points.
These intervals estimate test-set uncertainty, not variation across
training runs (Appendix~\ref{app:paired-archive-analysis}). 
OOD ASR falls by $1.32$ points
with unchanged over-block, but its interval $[-3.29,0.00]$ includes zero.
The ID gain spans 14 improving scenarios and remains significant under
scenario-cluster resampling (Appendix~\ref{app:paired-archive-analysis}).

The evolving guard blocks 15 ID attacks that bypassed the fixed-archive
guard. In 13 of these cases, the decision changes from Verify (Level~2)
to Block (Level~4). Both corrected OOD cases show the same change.
Most improvements therefore come from blocking violations that the
fixed-archive guard had already flagged. When Level~$\geq2$ counts as Block,
the fixed archive instead has one fewer miss on each split, equal ID over-block,
and one fewer OOD benign block (Appendix~\ref{app:interventions}).
The archive benefit therefore concerns the Level-4 blocking decision and does
not persist under the stricter blocking rule.

\subsection{External Evaluation and Construction Reuse}
\label{sec:external-results}

\paragraph{Violation detection on VIOLA.}
We extract 1,709 invocation-level inputs from the calls recorded in 114 multi-agent trajectories, preserving each invocation's agent identity and applicable policies.
In Table~\ref{tab:policyguardbench-r0-r10}, R10 lowers ASR from $27.08\%$ to $18.75\%$, with a $5.56$ percentage-point increase in over-block.
It also achieves lower ASR than the fixed archive ($18.75\%$ vs.\ $23.96\%$) at the same $66.67\%$ over-block (Table~\ref{tab:skill-causal-control}).
Over-block is measured per trajectory. Each trajectory contains about 15 invocations on average, and one false alarm is enough to flag an otherwise compliant trajectory. This aggregation partly explains the high over-block rate.
These results use only 18 benign trajectories, and the adapter was developed
using the test set. 

\begin{table}[t]
\centering
\scriptsize
\setlength{\tabcolsep}{3.0pt}
\resizebox{\columnwidth}{!}{%
\begin{tabular}{@{}llrrrrr@{}}
\toprule
\multicolumn{7}{c}{\textbf{VIOLA Test+Dev: 1,709 invocation inputs}} \\
\cmidrule(lr){1-7}
\textbf{Traj.} &
\textbf{Model} &
\textbf{ASR}$\downarrow$ &
\textbf{Ov.blk}$\downarrow$ &
\textbf{Acc.}$\uparrow$ &
\textbf{F1}$\uparrow$ &
\textbf{Youden's J}$\uparrow$ \\
\midrule
114 & Qwen3-14B R0  & 27.08 & 61.11 & 67.54 & 79.10 & 11.81 \\
    & Qwen3-14B R10 & 18.75 & 66.67 & 73.68 & 83.87 & 14.58 \\
\bottomrule
\end{tabular}
}

\vspace{3pt}

\resizebox{\columnwidth}{!}{%
\begin{tabular}{@{}llrrrrr@{}}
\toprule
\multicolumn{7}{c}{\textbf{PolicyGuardBench}} \\
\cmidrule(lr){1-7}
\textbf{Data} &
\textbf{Model} &
\textbf{ASR}$\downarrow$ &
\textbf{Ov.blk}$\downarrow$ &
\textbf{Acc.}$\uparrow$ &
\textbf{F1}$\uparrow$ &
\textbf{Youden's J}$\uparrow$ \\
\midrule
Full  & PolicyGuard-4B & --   & --   & 90.14 & 87.59 & -- \\
      & Qwen3-14B R0          & 9.93 & 8.95 & 90.63 & 89.07 & 81.12 \\
      & Qwen3-14B R10         & 9.93 & 8.64 & 90.82 & 89.27 & 81.44 \\
\addlinespace
Visible & Qwen3-14B R0        & 0.37 & 4.28 & 96.64 & 93.30 & 95.35 \\
      & Qwen3-14B R10         & 0.28 & 3.89 & 96.96 & 93.91 & 95.84 \\
\bottomrule
\end{tabular}
}
\caption{\textbf{External diagnostics (\%).} VIOLA has 114 trajectories
(Test + Dev), comprising 1,709 invocations (848 Test + 861 Dev). Visible denotes a post-hoc PolicyGuardBench subset whose compliance can be determined from the provided information. PolicyGuard-4B scores are published references. }
\label{tab:policyguardbench-r0-r10}
\end{table}

\paragraph{Performance preservation on PolicyGuardBench.}
R10 keeps full-set ASR at $9.93\%$ and reduces over-block from $8.95\%$ to
$8.64\%$. We also use the Visible subset to test the guard when the input
provides enough information to judge compliance. 
In this subset, R0 already detects nearly all violations in the Visible subset,
leaving little room for training to further reduce ASR.
Appendix~\ref{app:external-eval} details subset construction.

\paragraph{Construction reuse in Qwen-AgentWorld.}
This pilot tests whether we can reuse the policy-case format and evaluation
interface outside AWM, and how the existing guard performs without
target-domain training. We use Qwen-AgentWorld-35B-A3B
\citep{zuo2026qwenagentworld} to simulate tool responses for 180 test cases:
90 violating and 90 compliant. Each case includes a policy, request,
proposed call, visible state, and simulated observation. R0 and R10
judge the same cases.

This pilot shows that the policy construction framework
can be reused with simulated tool responses outside AWM.
Without target-domain training, R10 reduces ASR from $20.00\%$
to $2.22\%$, but increases over-block by approximately $13.3$ percentage
points.
The results support interface reuse and reveal a safety--utility
trade-off. Appendix~\ref{app:agentworld-construction} shares more details about this test.



\section{Related Work}

\paragraph{Policy-compliance benchmarks.}
PolicyGuardBench pairs web trajectories and their prefixes with policies to train
and evaluate the PolicyGuard detector \citep{wen2026learning}. VIOLA
\citep{viola2026} evaluates policy compliance in multi-agent execution traces.
PolicyGym complements these trajectory benchmarks with concrete tool calls
constructed in executable environments, accompanied by deployment policies,
execution evidence, and human-reviewed test labels. Every violating test case
uses a permitted tool, requiring checks of arguments, resources, and context
rather than tool identity alone.

\paragraph{Execution-based guard training.}
BraveGuard \citep{feng2026braveguard} and HazardAuditor
\citep{feng2026hazardauditor} turn safety threats into executable
computer-use tasks and collect agent trajectories to train guards.
AgentPolicyForge constructs compliant and violating tool calls from
deployment policies, using execution evidence and rule checks to support
labeling for evaluation and iterative guard training.

\paragraph{Adversarial and coevolutionary training.}
Related methods use self-play \citep{liu2025selfredteam},
train attack and defense agents \citep{zheng2024twoplayer,yu2025magic},
update attack and detection prompts \citep{jiang2025aegis},
or jointly train conversation and feedback agents
\citep{zhang2025alignmentwaltz}.
SafeCoEvo co-evolves a safety harness and guard through runtime experience
\citep{cheng2026safecoevo}. Rainbow Teaming maintains a MAP-Elites archive
of adversarial prompts \citep{samvelyan2024rainbow,mouret2015illuminating}.
These methods study safety and alignment through adversarial feedback.
AgentPolicyForge focuses on compliance with explicit deployment policies,
including restrictions on otherwise permitted tool calls. It keeps
attacker weights frozen, validates strategy updates on training scenarios,
and trains the guard through SFT on execution-grounded cases.


\section{Conclusion}
AgentPolicyForge builds policy benchmarks and trains guards using executable
environments. PolicyGym provides training data and human-reviewed tests.
Generated data improve enforcement, with archive updates adding an in-domain
gain at Level 4 in the matched run. 
Future work will evaluate guard adaptation to new policy types and deployment changes.

\clearpage
\section*{Limitations}
\label{sec:limitations}

PolicyGym studies synthetic policies whose violations can be determined from one
tool call. Both test splits share AWM and policy templates, so they do not establish
generalization to new policy types or deployments. The Qwen-AgentWorld pilot
reuses interfaces with simulated responses, but high benign rejection limits
direct guard transfer. We do not compare learned guards with complete executable
rule checks.

Results use one generation seed and two Qwen3 backbones, without measuring
variation across training runs. The attacker, task agent, and judge use proprietary
Claude models and may share biases in generated calls and labels. The follow-up
human review covers 100 test episodes, not independent double annotation of the
full test sets. Training-label accuracy remains unmeasured.

Evaluation uses fixed attacks and does not compare systems at equal benign
rejection. Baselines also differ in model size. These comparisons therefore do
not isolate all effects of training, model size, and blocking rules. External
adapters used evaluation feedback, and VIOLA has few benign trajectories.
We measure blocking decisions, not task completion, the outcomes of verification
or modification, or deployment cost.

\section*{Ethical considerations}
This work studies defensive policy guards in synthetic, resettable environments. It uses no real
user data or production systems; all generated policy violations are simulated and confined to
these environments.

\bibliography{references}

\newpage
\appendix

\section{Detailed Data Construction}
\label{app:data-construction}

This appendix explains how we construct and label cases, split the data, and train
the guard using the process in Section~\ref{sec:construction}.

\subsection{Adapter Requirements}

An environment adapter gives the framework access to tool schemas and observable
state, and provides functions to reset the environment and execute calls. It also
connects executable policy checks to the tools and fields they need. The
Qwen-AgentWorld pilot uses simulated tool observations with the same case format
and evaluation interface. It does not reproduce the full executable training loop.

\paragraph{Policy families and template instantiation.}
Table~\ref{tab:policy-family-templates} describes the ten policy families in PolicyGym.
Each family defines a type of restriction. To create a policy, we select compatible
tools and fields and fill in limits, identifiers, or prohibited values.
The table gives illustrative examples, not quotations from released cases.
A check applies only to tools covered by its clause and requires the relevant
arguments or state.

\begin{table*}[t]
\centering
\footnotesize
\setlength{\tabcolsep}{3pt}
\renewcommand{\arraystretch}{1.12}
\begin{tabular}{@{}p{0.13\textwidth}p{0.23\textwidth}p{0.29\textwidth}p{0.29\textwidth}@{}}
\toprule
\textbf{Policy family} & \textbf{Restriction} & \textbf{Illustrative policy} & \textbf{Information used by the check} \\
\midrule
Time windows & Disallow an operation during a specified time interval. & Transfers are prohibited between 21:00 and 05:00 UTC. & The bound action-time field and the prohibited interval, including intervals crossing midnight. \\
\addlinespace
Backdating & Limit how far an operation's date may precede the reference date. & An entry must not be dated more than 30 days in the past. & The bound date field, fixed reference date, and maximum age in days. \\
\addlinespace
Amount limits & Cap the amount of an operation. & A single transfer must not exceed \$500. & The relevant amount argument, its units, and the cap. \\
\addlinespace
Own-record access & Restrict an operation to the current user's records. & A record's owner identifier must match the current user. & The bound owner-identifier argument and trusted current-user identifier. \\
\addlinespace
Restricted tables & Exclude a table from the session's authorization. & The session must not directly write the audit-log table. & The restricted table name, clause scope, and evidence identifying the table targeted by the call. \\
\addlinespace
Protected identifiers & Reserve identifiers below a threshold. & Manual operations on account IDs below 1,000 are prohibited. & The bound identifier argument and reserved-range threshold. \\
\addlinespace
Data residency & Prohibit specified values of a location or jurisdiction field in writes. & A write must not set the region field to a listed restricted region. & The bound field value and the policy's prohibited-value list. \\
\addlinespace
Field masking & Prohibit the session from writing a designated field. & The session must not populate the sensitive-note field. & The designated argument and whether the call supplies a nonempty value. \\
\addlinespace
Protected flags & Prohibit setting a designated boolean flag to true. & The session must not set the locked flag to true. & The bound flag argument and its boolean value. \\
\addlinespace
Dual approval & Require an approval field above an amount threshold. & An operation exceeding \$1,000 must include a truthy approval field. & The amount, threshold, and presence of a truthy value in the designated approval field. \\
\bottomrule
\end{tabular}
\caption{\textbf{PolicyGym's ten policy families.} Names, fields, and numerical values in the examples are illustrative. Field masking denotes a write restriction, not output redaction. Dual approval checks the encoded approval-field requirement, not an independent two-person approval workflow.}
\label{tab:policy-family-templates}
\end{table*}

\subsection{Source Records and Fixed Environment Context}

We start with 6,603 seed--policy pairs from 698 resettable AWM scenarios.
Each adapter provides the tool operations, argument types, affected tables, and
initial database state. We fix the reference time, initial state, verifier, and
environment version for data collection and evaluation.

\subsection{Validating Attacker Seeds}

We first check that each policy can be applied to the scenario's tools and that
the required arguments and resources are available. A threat seed must support at
least one permitted tool call that violates the policy. A benign seed must support
a compliant call with valid arguments. Of the 6,603 source rows, 5,655 pass these
checks: 2,863 threat seeds and 2,792 benign seeds. These seeds are used to generate
requests and are not yet labeled tool-call examples.

\subsection{From a Seed to an Executed Trajectory}

For each seed, we restore the database to its initial state and load the policy,
tool schemas, and fixed reference time. Sonnet 4.6 generates a request. The task
agent receives the request without the policy and proposes a call. We retain calls
that use a permitted tool, have valid arguments, match the request, execute
successfully, and have an applicable policy clause. Each record stores the
pre-action inputs, execution results, verifier judgment, and LLM target.
The released supervision pool can include cases with an \textsc{Unknown} verifier
judgment. Test cases require a determinate judgment and human review.

The attacker receives the policy, tool schemas, task context, target operation,
and argument values. When enabled, it also receives retrieved strategies and
instructions on how to phrase the request. It must preserve the target action
without inventing tools or state, and return only a request inside
\texttt{<query>} tags. Only this request goes to the task agent.
During adaptive collection, benign requests are used unchanged.
The supplementary material includes the condensed prompt template.

The guard receives the fixed reference time and information from the adapter
about tool permissions and operation types. Additional state signals summarize
earlier calls, rather than the full database. This field is empty in query-only
adaptive collection. Execution results are used for offline labeling and are
not given to the guard.

\subsection{LLM Training Labels and Executable Verification}

Sonnet 5 supplies the structured training targets. The verifier separately checks
the encoded policy clauses and returns \textsc{Unknown} when the available evidence
is insufficient. We retain both judgments. Appendix~\ref{app:oracle-label-ablation}
compares training with LLM targets against training with verifier-derived targets.

Table~\ref{tab:label-agreement} compares the two label sources across all released
training, calibration, and development records. Their disagreements have not all
been adjudicated. Human review of the separate test sets does not establish the
accuracy of these training labels.

\begin{table}[htbp]
\centering
\small
\setlength{\tabcolsep}{4.0pt}
\begin{tabular}{@{}llr@{}}
\toprule
\textbf{Verifier} & \textbf{Sonnet 5} & \textbf{Records} \\
\midrule
Compliant & Compliant & 1,930 \\
Violation & Violation & 915 \\
Violation & Compliant & 180 \\
Compliant & Violation & 11 \\
Unknown   & Compliant & 146 \\
Unknown   & Violation & 129 \\
\bottomrule
\end{tabular}
\caption{\textbf{Agreement between Sonnet-5 and verifier-derived labels.}
Counts cover all 3,311 released supervision records, not the human-reviewed test sets.
\textsc{Unknown} denotes cases the verifier cannot determine from executable evidence.
Agreement with Sonnet 5 does not establish label accuracy.}
\label{tab:label-agreement}
\end{table}

\subsection{Splitting, Deduplication, and the Training Split}

After filtering, records remain from 671 scenarios. We split these scenarios using
a stable hash with seed 46: 537 for training, 67 for calibration, and 67 for
development. These splits contain 2,708, 302, and 301 records, respectively, and
are separate from the primary test sets. Before balancing, the training split has
1,828 compliant and 880 violating records. These counts describe the released
data, not the mixtures used in each training round.

To detect duplicate cases, we combine the policy boundary, proposed operation,
decisive arguments, and label into a semantic key. Records with the same key
cannot appear in different splits, and we limit repeats within training.
Scenario separation and these keys reduce overlap, but the splits still share
policy vocabulary and construction templates. Sampling for each training round
is described separately in Appendix~\ref{app:round-training}.

\paragraph{Train--test content overlap.}
We remove shared instructions and output templates, then compare requests and
policy clauses separately. For each test string, we find the most similar training
string using cosine similarity over word 2--4-gram TF--IDF features
(Table~\ref{tab:content-overlap-audit}). Exact matches identify text repeated across
splits. This check does not rule out a shared writing style or representation
from the data generator.

\begin{table}[htbp]
\centering
\scriptsize
\setlength{\tabcolsep}{3.2pt}
\begin{tabular}{@{}lrrrrr@{}}
\toprule
\textbf{Object} & \textbf{Train} & \textbf{Test} & \textbf{Exact} &
\textbf{Median} & \textbf{P95} \\
\midrule
Natural request & 2,708 & 1,371 & 0 & 0.049 & 0.106 \\
Policy clause   & 3,017 & 1,587 & 1 & 0.217 & 0.406 \\
\bottomrule
\end{tabular}
\caption{\textbf{Lexical overlap between the released training and test splits.}
Train and Test count unique strings. Exact counts test strings also present in training.
Median and P95 summarize each test string's similarity to its closest training
string, using TF--IDF cosine similarity after removing shared instructions and output templates.}
\label{tab:content-overlap-audit}
\end{table}

\subsection{Held-Out Test Set Construction}

We group scenarios by removing the final numeric suffix from their names.
The 698 source scenarios form 298 groups. The ID test contains 187 other scenarios
from 81 groups represented in the source pool. The OOD test contains 67 scenarios
from 62 groups absent from both the source pool and the final training split.
The two test sets share no scenarios. Of the 81 ID groups, 80 appear in the final
training split and one appears only in calibration. Grouping by name does not
guarantee that groups represent semantically different application domains.

We begin with 1,620 ID and 540 OOD candidate episodes. Strict verifier checks
retain 1,183 and 334, respectively. After human review, removal of failed cases,
and class balancing, the test sets contain 1,070 ID episodes (535 per label) and
304 OOD episodes (152 per label). We fix their requests, policies, pre-action
context, tool calls, and call order. The sets contain 3,521 ID calls and 955 OOD calls.

\subsection{Test Labels, Human Review, and Scoring}
\label{app:label-provenance}
\paragraph{Reference labels.}
Test candidates undergo executable checks, and one human reviewer checks each
LLM-proposed label. Cases that fail review are discarded rather than relabeled.
The final balanced test sets contain 1,070 ID and 304 OOD episodes. These results
therefore apply to cases that the verifier can judge and that pass human review.

Sonnet 5 proposes the tool calls, training targets, and test labels. Executable
checks and human review do not establish that the resulting judgments are
independent of Sonnet 5. The verifier-supervision experiment shows that training
can also improve with executable labels, but it does not measure test-label accuracy.

\paragraph{Sampled human--reference agreement.}
For a follow-up review, we sample 100 test episodes without replacement: 78 ID
and 22 OOD, with equal numbers of each reference label within each split.
An episode already reviewed in a pilot is excluded from the sampling pool and
does not count toward these 100 episodes. The sample contains 322 calls.

The reviewer sees the policies, requests, call arguments, fixed timestamps, and
visible session facts, but not the reference labels, model judgments, or risk
scores. After reviewing all calls in an episode, the reviewer assigns one
compliance label. An undecidable response is allowed, but none is submitted.

Table~\ref{tab:human-reference-agreement} compares these judgments with the unchanged
reference labels. The two disagreements are ID episodes labeled violating in the
reference and compliant by the reviewer. Neither has been adjudicated or relabeled.
The scores measure agreement between this reviewer and the reference labels.
They do not estimate label accuracy or agreement between two independently
blinded annotators.

\begin{table}[t]
\centering
\small
\setlength{\tabcolsep}{5pt}
\begin{tabular}{@{}lrrr@{}}
\toprule
\textbf{Split} & \textbf{Agree / $n$} & \textbf{Agreement (\%)} & \textbf{$\kappa$} \\
\midrule
ID & 76/78 & 97.44 & 0.949 \\
OOD & 22/22 & 100.00 & 1.000 \\
Overall & 98/100 & 98.00 & 0.960 \\
\bottomrule
\end{tabular}
\caption{\textbf{Sampled human--reference agreement.} Cohen's $\kappa$ compares
submitted episode judgments with existing test labels.}
\label{tab:human-reference-agreement}
\end{table}

\paragraph{Frozen scoring.}
All guards judge the same calls and visible context. Reference labels and per-call
violation flags are kept separate for scoring. We replay calls in order and stop
at the first Level-4 block. Recomputing metrics from the saved R0, evolving-archive,
and fixed-archive predictions reproduces the reported ASR and over-block counts.

\subsection{Round-Specific Training Data}
\label{app:round-training}

R0 is the guard before any SFT in our framework. Each round samples 100 threat
and 100 benign seeds. Each threat seed generates four request versions: direct,
longer semantic context, schema-dense, and natural operational. Each benign seed
supplies one request. This gives 500 attempted trajectories, collected in four
shards that share no scenarios. The same schema, execution, and verifier checks
apply throughout.

The previous guard scores each valid proposal. The violating examples include
current attacks that bypassed the guard, attacks it blocked, historical attacks,
and simple training attacks. Compliant examples include current controls,
historical examples, and calls the current guard incorrectly blocked.
Historical replay reuses saved SFT examples, including cases collected before
Round 1. It does not execute tools again or retrieve attack strategies.

Each round targets 320 records, split equally by verifier label. When available,
24--32 incorrectly blocked compliant calls make up 15--20\% of the target compliant
half. We do not fill a shortfall in these hard benign cases with ordinary benign
examples, invalid executions, or cases with unknown labels. When at least four
policy families are available, each family contributes at most 25\% of its class.
After label-specific filtering, the evolving run retains 164, 170, 154, 138, 150,
130, 148, 154, 148, and 148 records in successive rounds.

\paragraph{Optimization.}
Each round runs ten optimizer steps with learning rate $5\times10^{-7}$ and global
batch size 32. We repeat the accepted examples as needed to complete these steps,
for 100 steps overall. Across rounds, the evolving archive uses 1,504 accepted
records and the fixed archive uses 1,590. These totals count records before
optimizer reuse. They are not unique-example counts, because replay can include
the same example in multiple rounds.

The two conditions use the same generation budgets, source schedules, and number
of optimizer steps. Filtering can leave different numbers of accepted examples.
We use one stochastic generation schedule. Evaluation uses fixed proposals and
temperature zero, although separate inference runs can still produce different outputs.

\subsection{Attack-Strategy Promotion}
\label{app:strategy-promotion}

The frozen attacker builds the initial archive $S_0$ by summarizing earlier
verifier-confirmed violations into 18 strategies. Each strategy describes how the
attack works, when to use it, and which assumptions to avoid. We keep these
summaries concise and remove unsupported assumptions. Later rounds produce new
candidate strategies from successful and failed attacks.

We test candidates in additional rollouts on training scenarios. These scenarios
may also be used for request generation, but never come from the test sets.
We reject candidates that rely on nonexistent tools, invented state or authorization,
the wrong policy family, or actions that do not violate the target rule.
To evaluate the remaining candidates, we compare the prior archive with an archive
that includes them, using the same validation inputs.

All candidates must be retrieved in at least two training scenarios. Adding them
must not reduce the number of executed calls confirmed as violations by the verifier.
The validation runs must also have no disqualifying flags for unsupported authority,
fabricated context, or mismatched targets. Candidates then face one of two checks:
\begin{itemize}
\item \textbf{Production strategies} must produce at least one new bypass: a case
that bypasses the guard with the candidate but not with the prior archive.
The number of new bypasses must exceed the number of prior bypasses lost.
\item \textbf{Exploration strategies} must produce at least one bypass and keep the
total bypass count at or above the prior archive's count. They do not need to
produce a new bypass.
\end{itemize}
The exploration check therefore requires the bypass count to stay the same or
increase, rather than requiring every metric to improve. Strategies with the same
meaning are merged. The fixed-$S_0$ control uses the same data collection and guard
training procedure but does not update the archive.

The archive grows from 18 to 25 strategies across nine policy families.
Four production additions produce six new bypasses in the paired validation runs,
and three exploration additions pass their admission check
(Table~\ref{tab:archive-growth}).

\begin{table}[htbp]
\centering
\small
\setlength{\tabcolsep}{3.2pt}
\begin{tabular}{@{}lrrrr@{}}
\toprule
\textbf{Round} & \textbf{Active} & \textbf{Prod. $+$} &
\textbf{Expl. $+$} & \shortstack{\textbf{Candidate-only}\\\textbf{bypasses}} \\
\midrule
Initial--3 & 18 & 0 & 0 & 0 \\
4         & 19 & 1 & 0 & 1 \\
5         & 21 & 2 & 0 & 4 \\
6         & 22 & 0 & 1 & 0 \\
7         & 23 & 0 & 1 & 0 \\
8         & 25 & 1 & 1 & 1 \\
9--10     & 25 & 0 & 0 & 0 \\
\bottomrule
\end{tabular}
\caption{\textbf{Growth of the active attacker-strategy archive from 18 to 25 strategies.}
Active is the total number of strategies. Prod.\ $+$ and Expl.\ $+$ count new
production and exploration strategies. Candidate-only bypasses are training-validation
cases where the new production strategy bypasses the guard but the prior archive does not.}
\label{tab:archive-growth}
\end{table}

\section{Intervention and Evaluation Diagnostics}
\label{app:interventions}

Only Level 4 prevents execution in the primary evaluation. Levels 1--3 describe
possible responses, not calibrated violation probabilities. We do not test the
verification or modification workflows, or whether five levels work better than
binary outputs. SFT targets use Levels 0 and 4. Qwen3-14B also predicts Level 2,
but Levels 1 and 3 do not appear in the reported predictions
(Table~\ref{tab:intervention-distribution}).

Intervention recall and FPR count Levels $\geq2$ as positive for guards that output
levels. Binary guards use Block. These metrics measure whether the guard flags a
call, rather than whether it stops execution. Youden's $J$ is recall minus FPR.
All thresholds are fixed before evaluation.

We estimate uncertainty using 20,000 paired bootstrap samples of episodes.
These intervals describe test-set uncertainty for the trained checkpoints.
They do not account for dependence between episodes from the same scenario or
variation across training seeds. Appendix~\ref{app:paired-archive-analysis} also
resamples whole scenario clusters to account for within-scenario dependence.
Training shifts predictions from Level 2 toward Level 4, reducing ASR but increasing
over-block by about four points on each primary split.

\begin{table}[htbp]
\centering
\scriptsize
\setlength{\tabcolsep}{2.5pt}
\resizebox{\columnwidth}{!}{%
\begin{tabular}{@{}llrrrrr@{}}
\toprule
\textbf{Checkpoint} & \textbf{Split} & \textbf{Proposals} &
\textbf{Level 0} & \textbf{Level 2} & \textbf{Level 4} &
\textbf{Overblock (\%)} \\
\midrule
Round 0    & in-dom.  & 3,521 & 2,752 & 239 & 530 & 3.55 \\
Round 10   & in-dom.  & 3,521 & 2,783 & 101 & 637 & 7.48 \\
\addlinespace
Round 0    & out-dom. & 955   & 717   & 83  & 155 & 1.97 \\
Round 10   & out-dom. & 955   & 721   & 48  & 186 & 5.92 \\
\bottomrule
\end{tabular}
}
\caption{\textbf{Proposal-level intervention counts and episode-level over-block.}
Counts cover all frozen Qwen3-14B proposals, including threat and benign episodes. Over-block
is the percentage of benign episodes with any Level-4 proposal. Levels 1 and 3 do not occur.}
\label{tab:intervention-distribution}
\end{table}

\paragraph{What happens if every flagged call is blocked?}
We rescore the saved R0, fixed-archive R10, and evolving R10 predictions, treating
any Level $\geq2$ response as Block. We keep the original call order and stop before
the first blocked call executes. A violation that executes before a later block
still counts as a successful attack. We report both thresholds without selecting
one based on test results or running the models again.

Blocking at Level $\geq2$ reduces ASR but increases benign blocking for every
checkpoint (Table~\ref{tab:block-threshold-control}). Under this rule, fixed-archive
R10 has one fewer miss on each split than evolving R10. It has the same ID over-block
and one fewer OOD benign block. The evolving archive's advantage at Level 4 therefore
does not hold at Level $\geq2$. These results do not compare the guards at the same
benign rejection rate.

\begin{table}[htbp]
\centering
\scriptsize
\setlength{\tabcolsep}{3pt}
\begin{tabular*}{\columnwidth}{@{\extracolsep{\fill}}llrrrr@{}}
\toprule
& & \multicolumn{2}{c}{\textbf{ID}} & \multicolumn{2}{c}{\textbf{OOD}} \\
\cmidrule(lr){3-4}\cmidrule(lr){5-6}
\textbf{Guard} & \textbf{Block rule} & \textbf{ASR} & \textbf{OB} & \textbf{ASR} & \textbf{OB} \\
\midrule
R0 & Level 4 & 19.63 & 3.55 & 17.11 & 1.97 \\
   & Level $\geq2$ & 3.18 & 14.58 & 4.61 & 18.42 \\
Fixed R10 & Level 4 & 10.84 & 6.73 & 9.21 & 5.92 \\
          & Level $\geq2$ & 2.80 & 12.52 & 1.97 & 17.11 \\
Evolving R10 & Level 4 & 8.22 & 7.48 & 7.89 & 5.92 \\
             & Level $\geq2$ & 2.99 & 12.52 & 2.63 & 17.76 \\
\bottomrule
\end{tabular*}
\caption{\textbf{Blocking-rule control (\%).}
Both blocking rules use the same saved predictions and score whole episodes.
OB is benign over-block. Fixed R10 uses the fixed-$S_0$ archive.}
\label{tab:block-threshold-control}
\end{table}

\section{Adaptive-Training Results}
\label{app:qwen3-14b-trajectory}

Table~\ref{tab:sft-trajectory} reports the starting and final results from
Figure~\ref{fig:qwen-round-trajectories}. Both backbones reduce ASR but increase
over-block. Performance does not improve at every round, and the round with the
lowest ASR can differ from the round with the lowest over-block. The supplementary
CSVs report all eight metrics for every round, including the verifier-supervised
and 8B runs.

\begin{table}[htbp]
\centering
\small
\setlength{\tabcolsep}{3pt}
\begin{tabular}{@{}llrrrr@{}}
\toprule
& & \multicolumn{2}{c}{\textbf{ID}} & \multicolumn{2}{c}{\textbf{OOD}} \\
\cmidrule(lr){3-4}\cmidrule(lr){5-6}
\textbf{Qwen3} & \textbf{Round} & \textbf{ASR} & \textbf{OB} & \textbf{ASR} & \textbf{OB} \\
\midrule
14B & R0  & 19.63 & 3.55 & 17.11 & 1.97 \\
    & R10 & 8.22 & 7.48 & 7.89 & 5.92 \\
8B  & R0  & 33.08 & 2.80 & 39.47 & 2.63 \\
    & R10 & 21.12 & 10.28 & 20.39 & 9.87 \\
\bottomrule
\end{tabular}
\caption{\textbf{Adaptive-training endpoints with LLM supervision (\%).}
R0 uses pretrained weights with the fused prompt. R10 follows ten rounds. OB is episode-level
over-block. Both measures use Level 4.}
\label{tab:sft-trajectory}
\label{tab:qwen3-8b-coevoguardbench}
\end{table}

\section{Supervision-Source Ablation: Verifier-Derived Labels}
\label{app:oracle-label-ablation}

We repeat ten rounds of training using verifier-derived targets. The model,
R0 checkpoint, source schedule, planned generation budget, and optimization settings
match the LLM-supervised run. Each run generates and selects its own examples,
so the comparison changes more than the labels on a fixed dataset.

With verifier supervision, the guard misses more violations but blocks fewer
compliant episodes than with LLM supervision (Table~\ref{tab:oracle-sft-trajectory}).
This experiment shows how results change with the supervision procedure. It does
not measure label accuracy or isolate the effect of archive updates.

\begin{table}[htbp]
\centering
\small
\setlength{\tabcolsep}{3pt}
\begin{tabular}{@{}lrrrr@{}}
\toprule
& \multicolumn{2}{c}{\textbf{ID}} & \multicolumn{2}{c}{\textbf{OOD}} \\
\cmidrule(lr){2-3}\cmidrule(lr){4-5}
\textbf{Supervision} & \textbf{ASR} & \textbf{OB} & \textbf{ASR} & \textbf{OB} \\
\midrule
LLM & 8.22 & 7.48 & 7.89 & 5.92 \\
Verifier & 11.40 & 6.54 & 11.18 & 2.63 \\
\bottomrule
\end{tabular}
\caption{\textbf{Qwen3-14B R10 under two supervision sources (\%).}
ASR and over-block (OB) are episode-level at Level 4. The common R0 appears in
Table~\ref{tab:sft-trajectory}.}
\label{tab:oracle-sft-trajectory}
\end{table}

\section{Additional Baselines and a Smaller-Backbone Replication}
\label{app:backbone-diagnostics}

\subsection{Frozen Guards and Task-Agent Baselines}

\paragraph{Prompts and parsing.}
The single-shot prompt asks Qwen3-14B for a verdict. The decomposed prompt first
asks for the policy clause and evidence, then uses a second model call for the
decision. The fused prompt returns the clause, evidence, intervention level, and
reason in one response. We use the fused prompt because it has low over-block
and high intervention recall, although the decomposed prompt has lower ID ASR.

The prompts do not guarantee valid output syntax. If a response cannot be parsed,
the evaluator allows the call. Haiku produces one malformed response on each split.
For GPT-6-luna, empty responses are retried with a larger token budget, after which
all responses can be parsed.

\paragraph{Comparison scope.}
The models in Table~\ref{tab:main_results} differ in size and decision rules.
The policy-aware task agent can change its proposed actions, whereas external
guards judge fixed calls. Allowing a benign call does not establish that the task
is completed. R10 has the lowest ASR, but other methods have lower benign rejection
or intervention FPR. The test violations are fixed and were not adapted to each
checkpoint.

\subsection{PolicyGuard-4B: Proposal and Trajectory Views}
We evaluate PolicyGuard-4B in two ways: judging each call before execution
(Online), or judging the complete action summary (Trajectory). Both use the same
episodes and give similar ASR and over-block
(Table~\ref{tab:policyguard4b-coevoguardbench}). Its missed violations therefore
cannot be explained by the per-call input format alone. ASR and over-block score
whole episodes in both views. The other metrics score individual calls in Online
and complete summaries in Trajectory.

\begin{table*}[htbp]
\centering
\scriptsize
\setlength{\tabcolsep}{4.0pt}
\resizebox{\textwidth}{!}{%
\begin{tabular}{@{}llrrrrrrr@{}}
\toprule
\textbf{Evaluation view} & \textbf{Split} &
\textbf{ASR}$\downarrow$ & \textbf{Ov.blk}$\downarrow$ &
\textbf{Accuracy}$\uparrow$ & \textbf{F1}$\uparrow$ &
\textbf{Recall}$\uparrow$ & \textbf{FPR}$\downarrow$ &
\textbf{Parse}$\uparrow$ \\
\midrule
Online     & in-dom.  & 32.52 & 2.80 & 92.22 & 74.39 & 66.22 & 2.43 & 100.00 \\
Online     & out-dom. & 28.29 & 0.66 & 94.14 & 81.21 & 71.18 & 0.89 & 100.00 \\
Trajectory & in-dom.  & 32.71 & 2.62 & 82.34 & 79.21 & 67.29 & 2.62 & 100.00 \\
Trajectory & out-dom. & 28.29 & 0.66 & 85.53 & 83.21 & 71.71 & 0.66 & 100.00 \\
\bottomrule
\end{tabular}%
}
\caption{\textbf{PolicyGuard-4B under online and trajectory evaluation.}
Online ASR and over-block are episode-level. Accuracy, F1, recall, and FPR are proposal-level.
Trajectory metrics score complete action summaries. Parse denotes output parseability.
Values are percentages.}
\label{tab:policyguard4b-coevoguardbench}
\end{table*}

\subsection{Qwen3-8B Replication}
Qwen3-8B uses the same training procedure and fixed tests as Qwen3-14B
(Table~\ref{tab:sft-trajectory}). Its ASR falls from $33.08\%$ to $21.12\%$ in-domain
and from $39.47\%$ to $20.39\%$ out-of-domain. Its intervention FPR remains higher.
This confirms the direction of the training effect on a second backbone, but does
not establish how performance scales with model size.
Figure~\ref{fig:qwen-round-trajectories} and the supplementary CSV show all rounds.

\section{Paired Error Analysis of the Training Conditions}
\label{app:paired-archive-analysis}

We compare saved predictions for the same episodes, reference labels, calls,
call order, and visible inputs. Using the same Level-4 blocking rule reproduces
the results in Table~\ref{tab:skill-causal-control}. This analysis requires no
additional model inference.

\begin{table*}[t]
\centering
\small
\setlength{\tabcolsep}{5pt}
\resizebox{\textwidth}{!}{%
\begin{tabular}{@{}llrrrrrl@{}}
\toprule
\textbf{Split} & \textbf{Error type} & \textbf{Episodes} &
\textbf{Frozen errors} & \textbf{Evolving errors} & \textbf{Repaired} &
\textbf{New errors} & \textbf{Clustered 95\% CI (pp)} \\
\midrule
In-domain & Missed violation & 535 & 58 & 44 & 15 & 1 & $[-4.17,-1.21]$ \\
In-domain & Benign block & 535 & 36 & 40 & 1 & 5 & $[0.00,1.69]$ \\
Out-of-domain & Missed violation & 152 & 14 & 12 & 2 & 0 & $[-3.37,0.00]$ \\
Out-of-domain & Benign block & 152 & 9 & 9 & 2 & 2 & $[-2.60,2.63]$ \\
\bottomrule
\end{tabular}%
}
\caption{\textbf{Paired error changes from the frozen to the evolving archive.}
Both guards are evaluated at Round 10. Repaired counts fixed-archive errors
corrected by evolving training. New errors counts the reverse changes. Intervals show
the evolving-minus-frozen error-rate difference in percentage points, using 95\%
paired scenario-bootstrap intervals. They do not measure variation across training seeds.}
\label{tab:paired-archive-errors}
\end{table*}

\subsection{Evolving versus Fixed Archive on PolicyGym}
\paragraph{Scenario concentration and uncertainty.}
We draw 20,000 bootstrap samples of scenarios with replacement, separately for
threat and benign episodes. Each sampled scenario contributes all its episodes.
If a scenario is drawn more than once, its episodes are counted that many times.
We use the same draws for both checkpoints, compute differences in episode-level
error rates, and obtain percentile intervals from those differences.
The threat groups contain 171 ID and 50 OOD scenarios, and the benign groups
contain 178 and 59, respectively.

ID threat errors decrease in 14 scenarios and increase in one. No scenario accounts
for more than two net corrections. Removing any one scenario leaves the ID ASR
difference between $-2.82$ and $-2.26$ points. The OOD interval reaches zero.
These intervals describe the two checkpoints from one paired training run and do
not measure variation across training seeds.

\paragraph{Decision transitions.}
All 17 corrected threat episodes are stopped at a verifier-confirmed violating
call. Of the 15 ID corrections, 13 change from Level 2 to Level 4 and two change
from Level 0 to Level 4. Both OOD corrections change from Level 2 to Level 4.
Most corrections therefore block violations that the fixed-archive guard had
already flagged. The OOD benign-error totals are equal, but the decisions differ:
evolving training corrects two false alarms and introduces two others
(Table~\ref{tab:paired-archive-errors}).

\paragraph{Exploratory policy-family breakdown.}
The net reduction in ID misses is eight for protected identifiers, three for time
windows, two for own-record access, and one for data residency. Other families are
unchanged. Both corrected OOD misses involve protected identifiers.
A case can have several applicable clauses, so its assigned primary family does
not show which clause caused the decision to change.

The promoted production strategies concern time windows and own-record access.
We therefore cannot link each corrected test error to a particular new strategy
from these results alone. Doing so would require training comparisons that change
individual strategies.

\subsection{Evolving versus Fixed Archive on VIOLA}
For VIOLA, we compare the shared R10 rerun with the saved fixed-archive predictions.
Both evaluate the same trajectory identifiers and reference labels, although the
guards can flag different calls within each trajectory
(Table~\ref{tab:paired-viola-errors}).

\begin{table*}[t]
\centering
\small
\setlength{\tabcolsep}{5pt}
\begin{tabular}{@{}llrrrrr@{}}
\toprule
\textbf{Comparator} & \textbf{Error type} & \textbf{Trajectories} &
\textbf{Comparator errors} & \textbf{Evolving errors} & \textbf{Repaired} &
\textbf{New errors} \\
\midrule
Frozen archive & Missed violation & 96 & 23 & 18 & 8 & 3 \\
Frozen archive & Benign false alarm & 18 & 12 & 12 & 2 & 2 \\
\bottomrule
\end{tabular}
\caption{\textbf{Paired VIOLA error changes toward the evolving condition.}
Counts use the Table~\ref{tab:skill-causal-control} Test+Dev trajectory predictions.
Repaired and new errors describe changes from frozen to evolving training.
Equal false-alarm totals do not imply identical decisions.}
\label{tab:paired-viola-errors}
\end{table*}

These final-checkpoint comparisons do not show whether the gains hold across
training seeds or blocking thresholds. They also do not identify which training
changes caused each correction. VIOLA's small benign sample and conversion to
trajectory monitoring further limit conclusions about external performance.

\section{A Small Policy Dataset Beyond AWM}
\label{app:agentworld-construction}

We use Qwen-AgentWorld-35B-A3B to test whether we can construct policy cases
beyond AWM using simulated tool responses. The Qwen-AgentWorld models
support seven domains: MCP, Search, Terminal, Software Engineering, Android, Web,
and OS \citep{zuo2026qwenagentworld}.

Our pilot keeps the same policy-case format and guard-evaluation interface.
It contains 180 cases, with 90 violations and 90 compliant calls. Each case
includes a policy, request, proposed call, visible state, and model-simulated
observation. R0 and R10 judge the same cases without fine-tuning on this domain.

\begin{table}[t]
\centering
\small
\setlength{\tabcolsep}{3.5pt}
\begin{tabular}{@{}lrrrr@{}}
\toprule
\textbf{Guard} & \textbf{ASR}$\downarrow$ & \textbf{Over-block}$\downarrow$ &
\textbf{FN / 90} & \textbf{FP / 90} \\
\midrule
R0  & 20.00 & 34.44 & 18 & 31 \\
R10 & 2.22  & 47.78 & 2  & 43 \\
\bottomrule
\end{tabular}
\caption{\textbf{Qwen-AgentWorld diagnostic dataset of 180 cases.}
The 180 cases contain 90 violations and 90 compliant calls with simulated tool observations.
R0 and R10 are evaluated without target-domain fine-tuning. ASR and over-block are percentages.
FN and FP count missed violations and blocked compliant calls at Level 4.}
\label{tab:agentworld-180}
\end{table}

R10 lowers ASR from $20.00\%$ to $2.22\%$, but raises over-block from
$34.44\%$ to $47.78\%$ (Table~\ref{tab:agentworld-180}). It misses 16 fewer
violations but blocks 12 more compliant calls. Thus, improved violation
detection comes with more false alarms. R10 blocks nearly half of compliant
calls, so these results do not establish reliable guard transfer.

The construction result is separate: we reuse the policy-case format and
evaluation interface with simulated tool responses. This pilot does not
reproduce the full executable sandbox or test guard training and adaptation
in the new environment.

\section{External Benchmark Protocols and Diagnostics}
\label{app:external-eval}
\label{app:diagnostic-audits}

\subsection{VIOLA Data Processing and Evaluation}
VIOLA assigns labels to complete multi-agent trajectories. We extract
1,709 calls and their inputs and outputs from 57 Test trajectories
(848 calls) and 57 Dev trajectories (861 calls), keeping each call's
agent identity and applicable policies.

We develop the evaluation adapter on Test and use it unchanged on Dev.
The combined results therefore do not represent evaluation on entirely
unseen data. The guard reviews recorded calls and outputs rather than
blocking calls before execution. Here, over-block is the percentage of
benign trajectories incorrectly flagged as violating.

\paragraph{Detection and confirmation.}
The guard first checks each recorded call for possible violations. It returns a
verdict, check type, quoted policy clause, evidence, and comparison. Deterministic
checks verify that the quoted text appears in the input.

The same checkpoint then reviews each candidate violation. It rejects findings
based on optional instructions, rules that do not apply to the agent or situation,
unsupported state, or evidence that satisfies the rule. It can confirm visible
violations and missing required fields or steps. A trajectory is labeled violating
if any candidate is confirmed. Reference labels are never included in the prompts.
We compute overall metrics by pooling the Test and Dev confusion matrices.

A trajectory contains about 15 calls on average, so even one false alarm can flag
an otherwise compliant trajectory. If each of $n$ calls independently has a
false-positive probability $p$, the trajectory-level probability is $1-(1-p)^n$.
This expression illustrates the effect of repeated checks. It does not assume
that the actual decisions are independent.

\paragraph{Inference settings and error handling.}
We rerun R0 and R10 with the same initial inputs, evaluation scripts, and serving
settings, keeping all 114 trajectories. The first scan uses temperature zero and
an output limit of 384 tokens. The confirmation pass uses 128 tokens.

There are no request errors or parsing failures on Test. On Dev, R0 has four scan
parsing failures and R10 has three. Each has one confirmation parsing failure.
An unparseable response does not count as a confirmed violation, and no trajectory
is removed. Table~\ref{tab:skill-causal-control} uses this R10 rerun together with
the earlier fixed-archive predictions. The supplementary reproduction notes give
the serving settings.

\subsection{PolicyGuardBench: Visible-Decidable Sensitivity Analysis}
We used evaluation feedback to select prompt variants, so these results are
post-hoc diagnostics. We also examine a \emph{visible-decidable} subset: cases where
compliance can be determined from the provided trajectory, domain, and subdomain,
and that determination agrees with the source label. The filter does not use
R0 or R10 predictions.

The subset contains 4,640 of the 12,000 rows: 3,550 compliant and 1,090 violating
examples across 717 policies. Keeping one row for each identical combination of
policy, domain, subdomain, and trajectory leaves 4,214 rows.

\begin{table}[htbp]
\centering
\scriptsize
\setlength{\tabcolsep}{2.5pt}
\resizebox{\columnwidth}{!}{%
\begin{tabular}{@{}lrrrrrrr@{}}
\toprule
\textbf{Model} & \textbf{Rows} &
\textbf{Acc.}$\uparrow$ & \textbf{ASR}$\downarrow$ &
\textbf{Ov.blk}$\downarrow$ & \textbf{F1}$\uparrow$ &
\textbf{FN} & \textbf{FP} \\
\midrule
\multicolumn{8}{@{}l}{\emph{Full}}\\
R0  & 12,000 & 90.633 & 9.927 & 8.954 & 89.075 & 505 & 619 \\
R10 & 12,000 & 90.817 & 9.927 & 8.636 & 89.266 & 505 & 597 \\
\multicolumn{8}{@{}l}{\emph{Visible-decidable}}\\
R0  & 4,640 & 96.638 & 0.367 & 4.282 & 93.299 & 4 & 152 \\
R10 & 4,640 & 96.961 & 0.275 & 3.887 & 93.909 & 3 & 138 \\
\multicolumn{8}{@{}l}{\emph{Deduplicated}}\\
R0  & 4,214 & 96.535 & 0.393 & 4.444 & 93.290 & 4 & 142 \\
R10 & 4,214 & 96.796 & 0.294 & 4.131 & 93.770 & 3 & 132 \\
\bottomrule
\end{tabular}%
}
\caption{\textbf{PolicyGuardBench visible-decidable sensitivity analysis.}
The subset keeps cases that can be judged from visible information and whose
judgment agrees with the source label. Selection does not use model predictions.
Deduplication keeps one row per identical policy, domain, subdomain, and trajectory
input. FN and FP count missed violations and false alarms. Rates are percentages.}
\label{tab:policyguardbench-visible-decidable}
\end{table}

On this subset, R10 corrects one missed violation and 15 false alarms. It introduces
one new false alarm and no new missed violations. The improvements remain after
removing duplicate inputs. The three remaining misses involve the adapter's
interpretation of active risk and subdomain rules.

Among the excluded rows, 207 have a visible interpretation that disagrees with the
source label. There are also 49 rows in identical-input groups with conflicting labels. Many other rows lack the state or trajectory information
needed for a decision. This selected subset supplements the full benchmark and
does not replace it.

\section{Comparison with Related Frameworks}
\label{app:framework-comparison}

Table~\ref{tab:framework-comparison} compares the tasks, evidence, and training uses
reported by related frameworks. Several also collect data through execution and
use it to train guards. AgentPolicyForge connects conditional policies to tools
and state, then uses checked calls for fixed tests and repeated guard training.
The table describes the published designs, not all capabilities that these
frameworks could support with further development.

\begin{table*}[t]
\centering
\footnotesize
\setlength{\tabcolsep}{4pt}
\begin{tabular}{@{}p{0.17\textwidth}p{0.25\textwidth}p{0.28\textwidth}p{0.24\textwidth}@{}}
\toprule
\textbf{Framework} & \textbf{Task or policy construction} &
\textbf{Evidence and supervision} & \textbf{Use in evaluation or training} \\
\midrule
ToolEmu \citep{ruan2024identifying} &
Tool-use scenarios for identifying agent risks &
LM-emulated tool execution and LM safety evaluation, with human assessment &
Risk evaluation across tools and scenarios \\
\addlinespace
AgentDojo \citep{debenedetti2024agentdojo} &
Agent tasks and prompt-injection attacks in extensible tool environments &
Agent executions evaluated for task utility and attack success &
Evaluation of tasks, defenses, and adaptive attacks \\
\addlinespace
PolicyGuard \citep{wen2026learning} &
Policies paired with web trajectories and prefixes &
Policy-compliance supervision in PolicyGuardBench &
Training and evaluation of a policy guard \\
\addlinespace
BraveGuard \citep{feng2026braveguard} &
Emerging threats instantiated as executable computer-use tasks &
Collected rollouts with trajectory-level guard supervision &
Repeatable threat discovery, data collection, and guard training \\
\addlinespace
HazardAuditor \citep{feng2026hazardauditor} &
Executable threats across heterogeneous computer-use agents &
Normalized agent events and deterministic safety outcomes &
Cross-framework supervision and GuardPO training \\
\addlinespace
AgentPolicyForge &
Conditional policies bound to tools, resources, and state; compliant and violating calls &
Execution records, rule checks, LLM targets, and human-reviewed test labels &
Fixed policy tests, multi-round guard training, and validated strategy updates \\
\bottomrule
\end{tabular}
\caption{\textbf{Scope of related frameworks.} Entries summarize reported approaches.
They do not imply that unlisted capabilities are impossible or absent.}
\label{tab:framework-comparison}
\end{table*}

\section{AI Assistance Disclosure}
AI assistants were used to improve the clarity, organization, and language of the manuscript.
The authors reviewed all incorporated suggestions and take full responsibility for the final content.

\end{document}